\documentclass[]{spie}  %>>> use for US letter paper
\usepackage{amsmath,amsfonts,amssymb}
\usepackage{graphicx}
\usepackage{subcaption}
\usepackage[colorlinks=true, allcolors=blue]{hyperref}
\newcommand{\Cov}{\mathrm{Cov}}

\title{Deep learning denoiser assisted roughness measurements extraction from thin resists with low Signal-to-Noise Ratio (SNR) SEM images: analysis with SMILE} 

\author[a,b]{Sara Sacchi}
\author[a]{Bappaditya Dey}
\author[c]{Iacopo Mochi}
\author[a]{Sandip Halder}
\author[a]{Philippe Leray}
\affil[a]{imec, Kapeldreef 75, 3001 Leuven, Belgium}
\affil[b]{Department of Physics and Astronomy, University of Bologna, 40126 Bologna, Italy}
\affil[c]{Paul Scherrer Institut, 5303 Villigen, Switzerland}

\authorinfo{Further author information: (Send correspondence to Bappaditya Dey)\\Bappaditya Dey: E-mail: Bappaditya.Dey@imec.be}

\begin{document} 
\maketitle

\begin{abstract}
The technological advance of High Numerical Aperture Extreme Ultraviolet Lithography (High NA EUVL) has opened the gates to extensive researches on thinner photoresists (below 30$\sim$nm), necessary for the industrial implementation of High NA EUVL. Consequently, images from Scanning Electron Microscopy (SEM) suffer from reduced imaging contrast and low Signal-to-Noise Ratio (SNR), impacting the measurement of unbiased Line Edge Roughness (uLER) and Line Width Roughness (uLWR). Thus, the aim of this work is to enhance the SNR of SEM images by using a Deep Learning denoiser and enable robust roughness extraction of the thin resist. For this study, we acquired SEM images of Line-Space (L/S) patterns with a Chemically Amplified Resist (CAR) with different thicknesses (15$\sim$nm, 20$\sim$nm, 25$\sim$nm, 30$\sim$nm), underlayers (Spin-On-Glass - SOG, Organic Underlayer - OUL) and frames of averaging (4, 8, 16, 32, and 64$\sim$Fr).
After denoising, a systematic analysis has been carried out on both noisy and denoised images using an open-source metrology software, SMILE 2.3.2, for investigating mean CD, SNR improvement factor, biased and unbiased LWR/LER Power Spectral Density (PSD). 
Denoised images with lower number of frames present unaltered Critical Dimensions (CDs), enhanced SNR (especially for low number of integration frames), and accurate measurements of uLER and uLWR, with the same accuracy as for noisy images with a consistent higher number of frames.
Therefore, images with a small number of integration frames and with SNR$\sim$$< 2$ can be successfully denoised, and advantageously used in improving metrology throughput while maintaining reliable roughness measurements for the thin resist.
%LIKE THIS IS 250 WORDS
\end{abstract}
%ABSTRACT MAX 250
%KEYWORDS MAX 8

% Include a list of keywords after the abstract 

\section{INTRODUCTION}
\label{sec:intro}
Moore's Law, stating that transistors density in electronic devices approximately doubles every two years, has been sustained until today thanks to continuous multi-directional innovations, such as Extreme Ultraviolet Lithography (EUVL), with wavelenght of $\lambda = 13.5$$\sim$nm. EUVL has gained popularity over the past few decades and, currently,is being used for high volume manufacturing of semiconductor devices worldwide. It is one of the most efficient ways of scaling the smallest size of the features printed on the photo-resist of the silicon chip, which is called critical dimension (CD). Its scaling is described in Eq. \ref{eq3}:
\begin{equation}
    CD=k_{1}\frac{\lambda}{NA}
\label{eq3}
\end{equation}
where $k_1$ is a factor describing the ability of the lithographic process to resolve small features, $\lambda$ is the wavelength of the light source and $NA$ is the lens numerical aperture used to pattern the silicon chip.

Recently, High Numerical Aperture EUV Lithography - High-NA EUVL ($NA = 0.55$, instead of $0.33$ for EUVL) has attracted great interest and a lot of research is being carried out to explore this technique, which  holds immense potential for industrial applications and will enable further scaling of Integrated Circuits (ICs). Towards employing High-NA EUVL for production, one of the major requirements to be satisfied is thinner photoresists to avoid pattern collapse. Thin resist (with thickness below $30 \sim nm$), as well as different resist-underlayer combinations (CAR/MOR, SOG/OUL, etc.), have posed significant challenges towards robust and accurate roughness measurement extraction and analysis due to low imaging contrast and Signal-to-Noise Ratio (SNR). In the field of signal processing, SNR is used to estimate the level of noise present in a recorded signal \cite{bose2003digital} and it is defined as in Eq. \ref{eq4}:
\begin{equation}
    SNR=10\log_{10}{\frac{\sigma^2_{signal}}{\sigma^2_{noise}}} \hspace{2cm} [dB]
\label{eq4}
\end{equation}

In the Scanning Electron Microscope (SEM) image formation process, each step in signal generation and processing is a source of noise, which increases the complexity of the final image \cite{marturi2014scanning}. Recent studies \cite{https://doi.org/10.1002/sca.20282} have described the main noise contributions as primary emission, secondary emission, scintillator, photocathode, and photomultiplier, showing that every noise contribution is given by quantum fluctuations and follows a Poisson statistics. 
An important parameter affecting noise is the electron dose. For a Poisson distribution, the number of detected electrons is proportional to the number of electrons impinging on the sample, therefore increasing the electron dose would decrease the noise, but may lead to sample damage for photoresists (e.g., resist slimming or shrinkage) \cite{mack2018reducing}. The linear correlation with the frames of integration makes it possible to modulate the electron dose per pixel, fixing it as low as the final image noise allows.

\begin{figure} [ht]
\begin{center}
\begin{tabular}{c} %% tabular useful for creating an array of images 
\includegraphics[height=2.5cm]{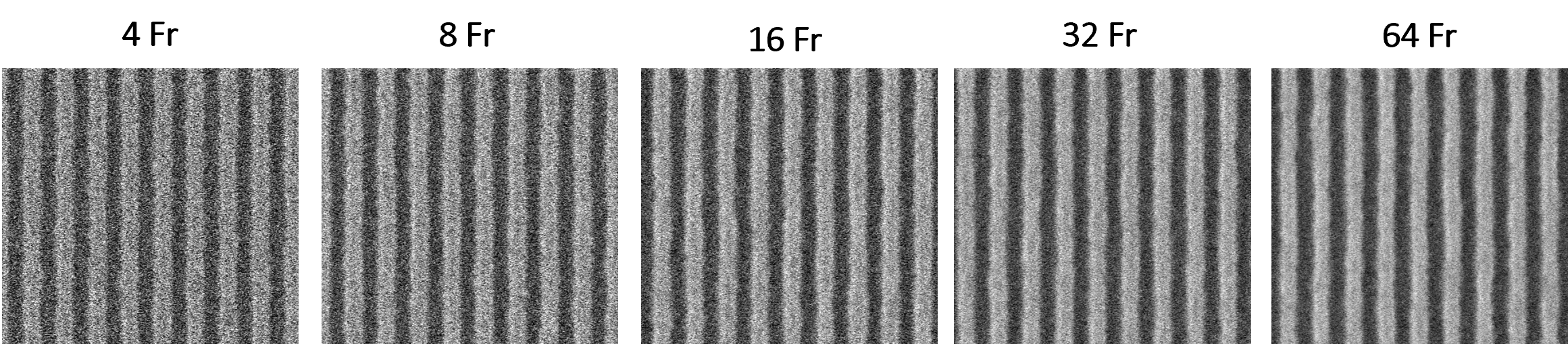}
\end{tabular}
\end{center}
\caption
%>>>> use \label inside caption to get Fig. number with \ref{}
{ \label{fig1} Example of SEM images with different frames of integration (4 Fr, 8 Fr, 16 Fr, 32 Fr, and 64 Fr).}
\end{figure}

In a nutshell, for photoresists, CD-SEM measurements with a large number of frames result in less noise, but require more time and can't completely mitigate the resist shrinkage effect; as opposed to a small number of frames, which leads to a higher edge detection uncertainty and unreliable roughness measurements (Fig. \ref{fig1}).

\begin{figure} [ht]
\begin{center}
\begin{tabular}{c} %% tabular useful for creating an array of images 
\includegraphics[height=5cm]{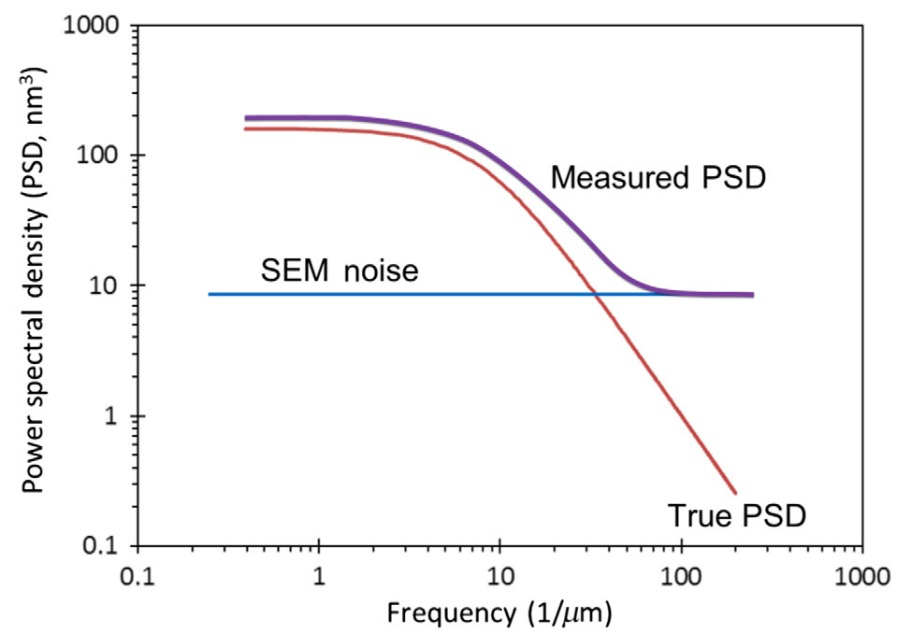}
\end{tabular}
\end{center}
\caption
%>>>> use \label inside caption to get Fig. number with \ref{}
{ \label{fig2} Example of noise subtraction from the measured PSD: the variance of the noise is subtracted from the biased variance, which is the area under the curve, to get the true unbiased PSD \cite{mack2018reducing}.}
\end{figure}

Line Edge Roughness (LER) and Line Width Roughness (LWR) are essential metrics for line/space patternes and they can conveniently be extracted from the Power Spectral Density (PSD) of the edges profiles. The PSD is the variance of the edge/width per unit frequency and it is calculated as the squared magnitude of the Fourier transform coefficients of the edge/width deviations \cite{mack2013systematic}. The area under the PSD curve is the variance of the roughness $\sigma^2_{biased}$, which is biased by the noise from the CD-SEM, as described in Eq. \ref{eq5}:
\begin{equation}
    \sigma^2_{biased} = \sigma^2_{unbiased} + \sigma^2_{noise}
\label{eq5}
\end{equation}
where $\sigma^2_{biased}$ is the roughness measured by the CD-SEM, $\sigma^2_{unbiased}$ the true variance of the wafer and $\sigma^2_{noise}$ the random SEM error in edge/width position (Fig. \ref{fig2}) \cite{mack2018reducing}. 

Due to the strongly reduced image contrast and poor SNR, the uncertainty in the contour extraction algorithms used for roughness and CD analysis with CD-SEM has greatly increased. More specifically, the increased noise level of thin resist images has a direct impact on the critical measurements of LER and LWR.
Recent studies \cite{10.1117/1.JMM.21.2.021207} have indicated that, if the SNR drops under a certain boundary value ($< 2.0$), the unbiased Line Width Roughness (uLWR) measurement may not reach the asymptotic plateau and becomes unreliable. In this case, for small pitch values, the measured edge position $w_{meas}$ depends on the true edge position $w_{true}$ plus an error $e$, such as $w_{meas} = w_{true} + e$ \cite{mack2022unbiased}. Thus, Eq. \ref{eq5} becomes:
\begin{equation}
    \sigma^2_{biased} = \sigma^2_{unbiased} + \sigma^2_{noise} + 2\Cov(w_{true},e)
\label{eq6}
\end{equation}
where $\Cov(w_{true},e)$ is the covariance of $w_{true}$ and $e$.

Therefore, to obtain accurate uLWR, higher SNR is imperative by generally acquiring and averaging larger number of frames. However, this requirement leads to extensive measurements time and reduced throughput on the current CD-SEM tools. An ideal solution would be to extensively measure images with smaller number of frames, which requires less time, and to efficiently denoise the raw images, in order to increase their SNR and extract accurate uLWR measurements. The most efficient and accurate tool for this task is Deep Learning.

\section{EXPERIMENTAL DESIGN AND METHODS}
\begin{figure} [h!]
\begin{center}
\begin{tabular}{c} %% tabular useful for creating an array of images 
\includegraphics[height=6cm]{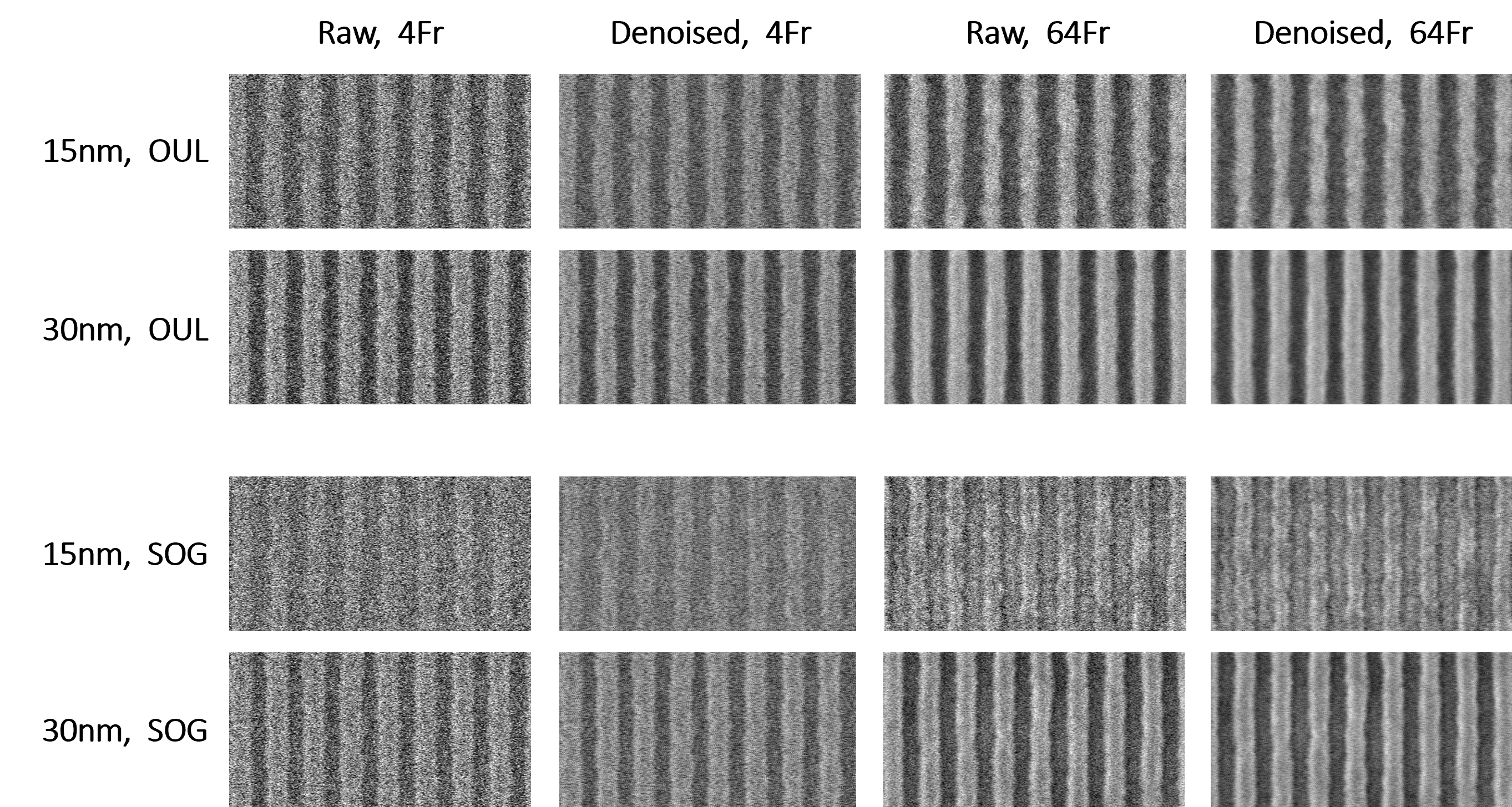}
\end{tabular}
\end{center}
\caption
%>>>> use \label inside caption to get Fig. number with \ref{}
[Example of noisy and denoised SEM images.]{ \label{exp1} Example of both noisy and denoised SEM images, acquired using different underlayers (SOG, OUL), resist thickness (15$\sim$nm, 30$\sim$nm), and frames of integration (4 Fr, 64 Fr).}
\end{figure}

As described at the end of Sec. \ref{sec:intro}, Deep Learning (DL) can be used to improve the uLWR measurements of CD-SEM images patterned through High NA EUVL. In fact, when using a smaller number of frames per image in the SEM tool in order to do extensive measurements in less time, the resulting images present low SNR, which affects negatively the unbiased LWR/LER. Thus, in this work a DL approach to improve the SNR of CD-SEM images is presented, which is able to enhance the SNR without changing the electron dose/frames. The model we used is a previously developed U-Net architecture-based unsupervised denoiser \cite{dey2021sem}.
The experimental dataset, on which the denoiser has been trained and tested, consists of SEM images captured with Hitachi High Tech CD-SEM CG6300 of a Chemically Amplified Resists (CAR) with two distinct underlayers, Spin-On-Glass (SOG) and Organic Underlayer (OUL), and four different thicknesses for the thin film resist: 15$\sim$nm, 20$\sim$nm, 25$\sim$nm, and 30$\sim$nm. The images have been acquired with increasing frames of integration (4$\sim$Fr, 8$\sim$Fr, 16$\sim$Fr, 32$\sim$Fr, and 64$\sim$Fr) with e-beam setting of 8$\sim$pA at 500$\sim$eV of energy, resulting in 50 images of $2048\times2048$ pixels with size of 0.8$\sim$nm per configuration.
In Fig. \ref{exp1} an example of raw and denoised images per configuration is depicted.

\begin{figure} [h!]
\begin{center}
\begin{tabular}{c} %% tabular useful for creating an array of images 
\includegraphics[height=6cm]{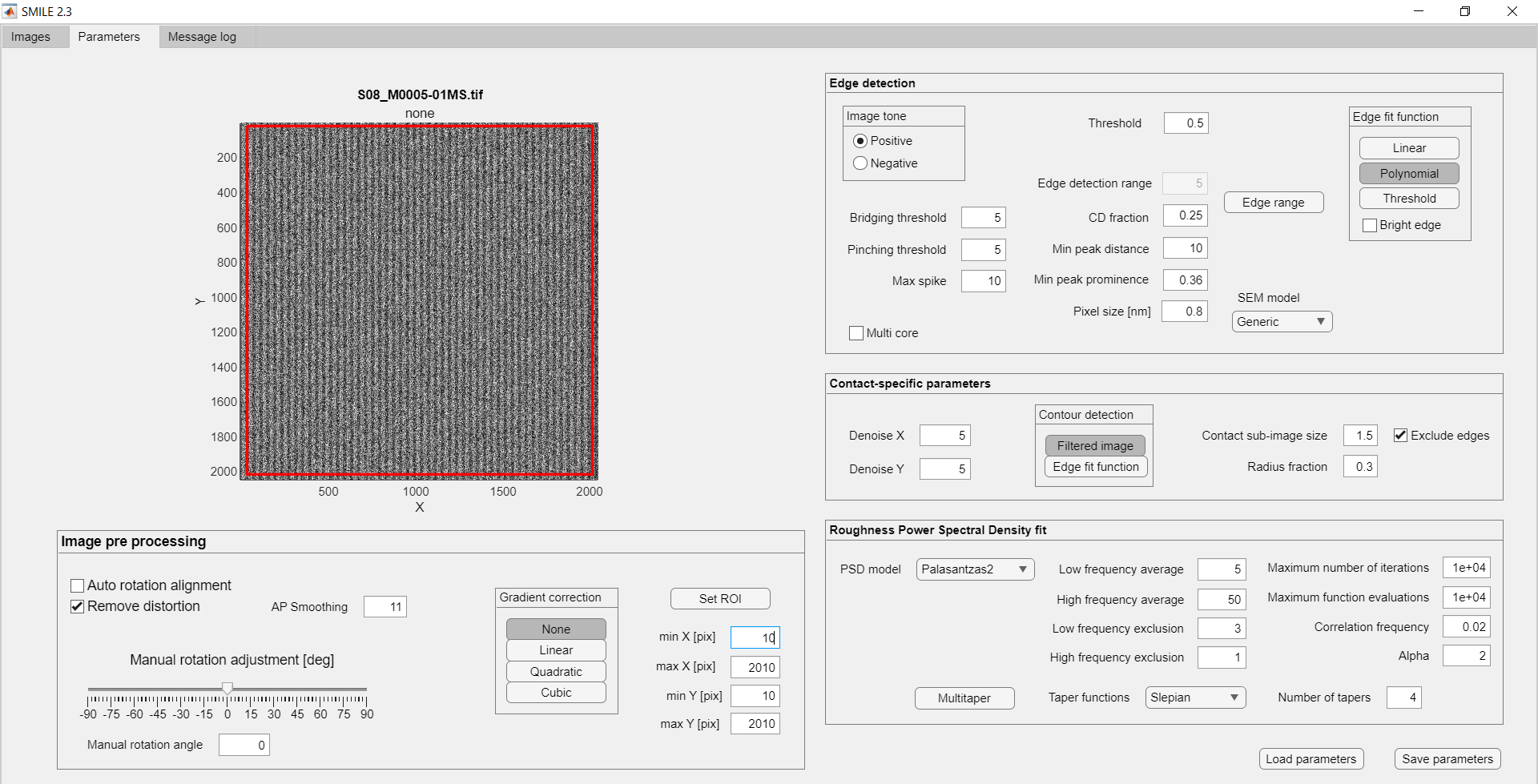}
\end{tabular}
\end{center}
\caption
%>>>> use \label inside caption to get Fig. number with \ref{}
[Example of noisy and denoised SEM images.]{ \label{smile} Example of SMILE 2.3 parameters panel.}
\end{figure}

After denoising, a systematic analysis has been carried out on both raw (noisy) and denoised images using an open-source metrology software, SMILE 2.3 \cite{mochi2021contacts}. The user interface for SMILE 2.3 is shown in Fig. \ref{smile}. For the experimental analysis, the parameters have been tuned by changing the default values in the following way:
\begin{itemize}
    \item Auto rotation alignment: off;
    \item Pixel size [nm]: $0.8$;
    \item Edge fit function: \textit{Polynomial};
    \item PSD model: \textit{palasantzas1} and \textit{palasantzas2};
    \item Low frequency exclusion: $3$.
\end{itemize}

\section{RESULTS AND DISCUSSION}
A set of experiments and analysis of SEM images with different combinations of underlayer, thickness and frames of integration, has been carried out to study how the linescan SNR, the mean CD and the biased and unbiased LWR PSD change between noisy and denoised images. The final goal of the experiment is to assess how the denoising procedure, using Machine Learning models, affect the SEM images. In particular, it is important to probe that such a denoising method don't have repercussions on the characteristic features of the device (e.g., the mean CD between noisy and denoised images should not be changed), but decreases the noise of the images (e.g., increasing the linescan SNR and decreasing the noise in the LWR PSD). This research proves to be advantageous towards initiating the first step for the employment of High NA EUVL, an innovative tehcnique, for HVM.

\subsection{Signal-to-Noise Ratio (SNR)}
\begin{figure} [h!]
\begin{center}
\begin{tabular}{c} %% tabular useful for creating an array of images 
\includegraphics[height=6cm]{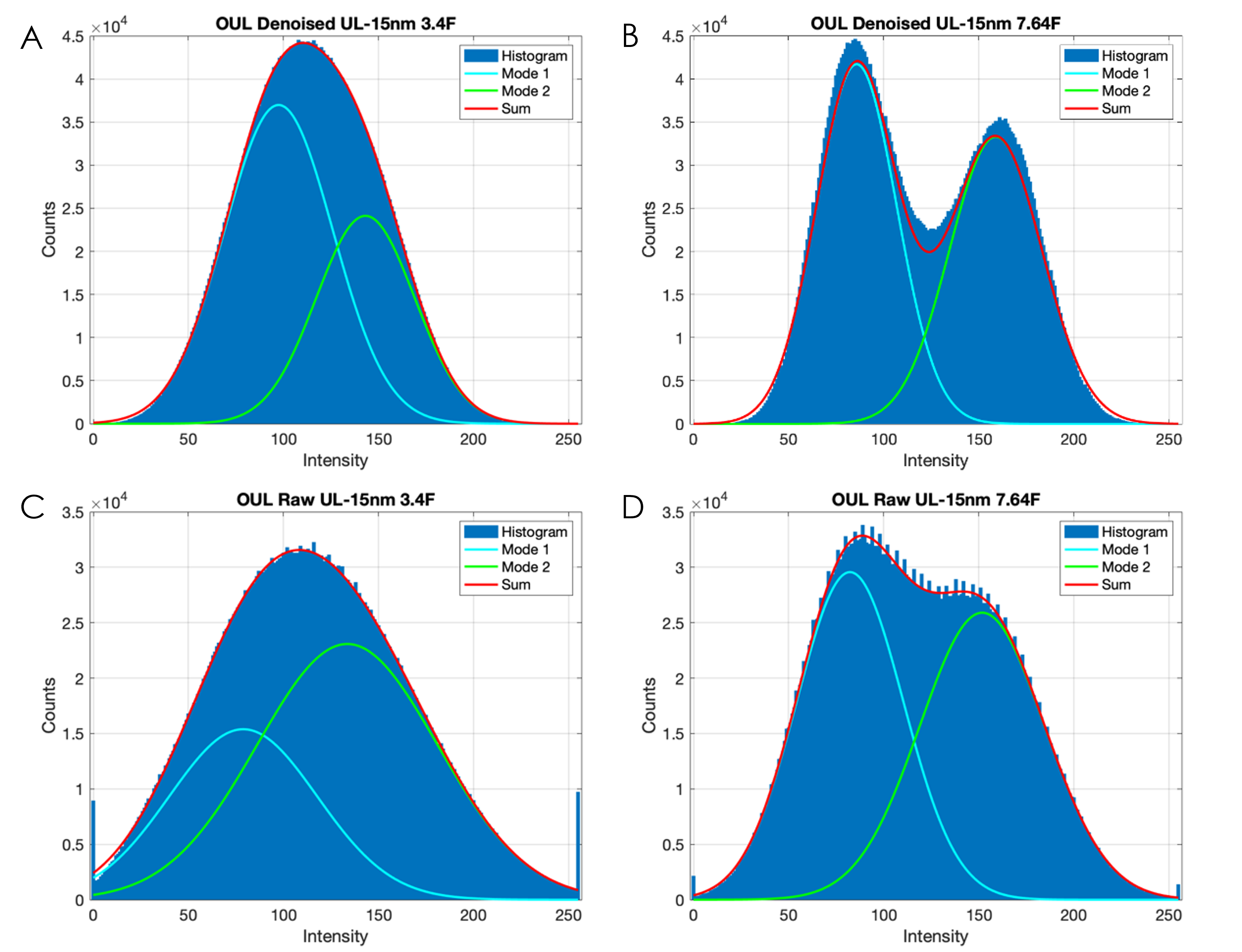}
\end{tabular}
\end{center}
\caption
%>>>> use \label inside caption to get Fig. number with \ref{}
[Example of grayscale distributions of SEM images.]{ \label{grayscale} Example of linescan SNR computation procedure from grayscale distribution of denoised SEM images acquired with 4 and 64 frames of integration.}
\end{figure}

For each image of the different configurations a histogram of the grayscale values has been plotted and fitted using two gaussians distributions for the maximum and minimum intensity, as described in Sec. \ref{sec:intro}. 
After the denoising procedure, the bi-modal nature of the grayscale distribution becomes apparenet.
Linescan SNR values are extrapolated from the gaussian distributions by fitting two gaussian peaks to the grayscale. Fig. \ref{grayscale} shows an example of the procedure of extrapolating the SNR from two fitted normal distributions, assumed for both intensity values, so that the histogram of the image can be represented with the model in Eq. \ref{H}:
\begin{equation}
    H(I)=M_1e^{-2\frac{I-I_1}{2\sigma_1^2}}+M_2e^{-2\frac{I-I_2}{2\sigma_2^2}}
\label{H}
\end{equation}
where $M_1$ and $M_2$ are the normal distribution parameters, $I_1$ and $I_2$ are the intensities of the pixels in the space region and in the line regions respectively, and $\sigma_1$ and $\sigma_2$ are the corresponding standard deviations.
To estimate the SNR for these images this model is fitted to the histogram of each image, and the SNR is calculated as the difference between the intensity values divided by the average standard deviation, as in Eq. \ref{SNR1}:
\begin{equation}
    SNR=2\frac{I_1-I_2}{\sigma_1+\sigma_2}
    \label{SNR1}
\end{equation}

\begin{figure} [h!]
\begin{center}
\begin{tabular}{c} %% tabular useful for creating an array of images 
\includegraphics[height=6cm]{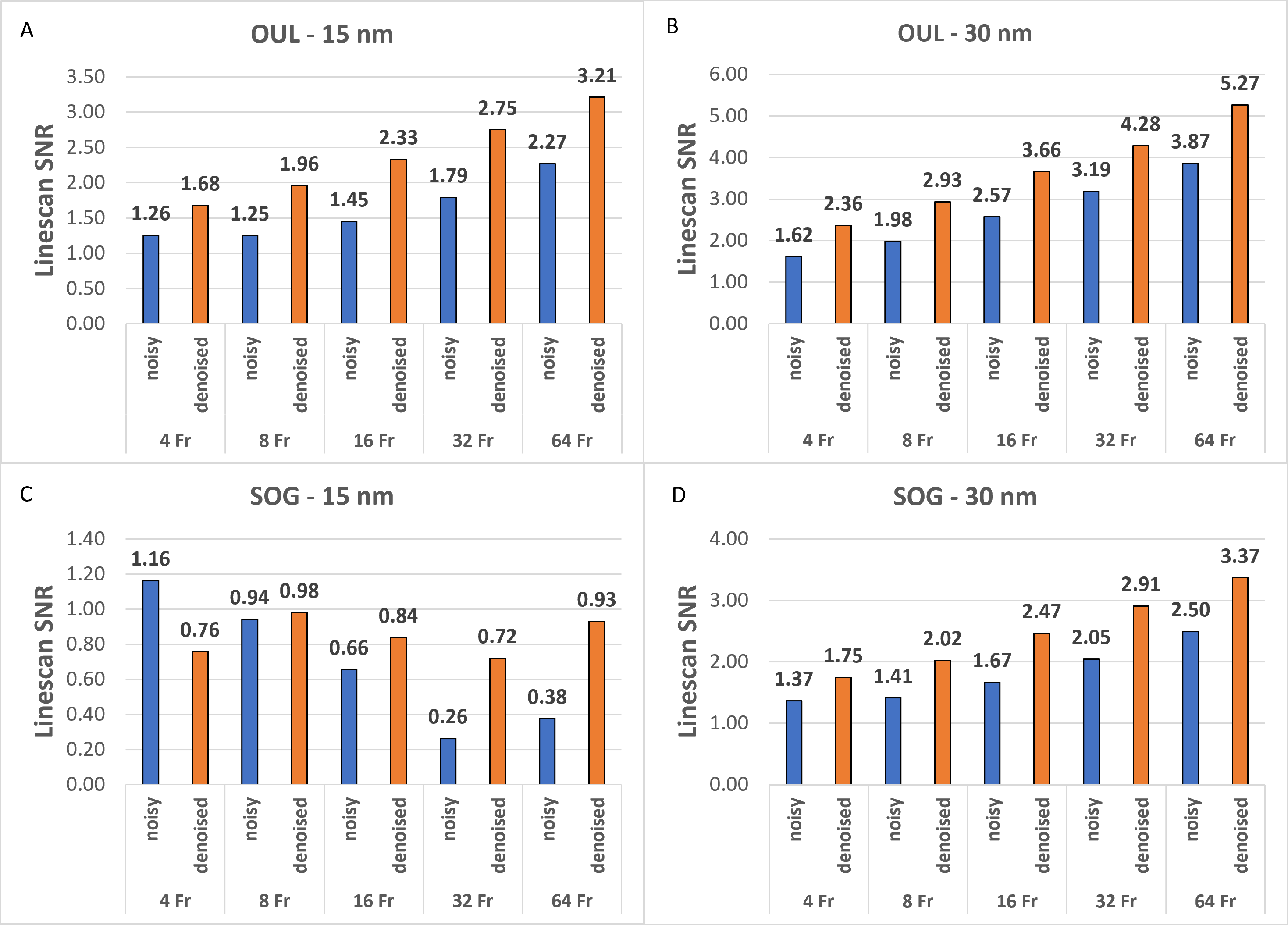}
\end{tabular}
\end{center}
\caption[Linescan SNR values before and after denoising.]
%>>>> use \label inside caption to get Fig. number with \ref{}
{ \label{snr} Linescan SNR values before and after denoising procedure of SEM images of different underlayers (OUL, SOG), resist thicknesses (15 and 30$\sim$nm), and frames of integration (4, 8, 16, 32, 64$\sim$Fr).}
\end{figure}

In Fig. \ref{snr} the SNR values of noisy and denoised images are compared through an histogram, while in Table \ref{tab:snr} the relative percentage difference between noisy and denoised SNR values are reported in percentage of the noisy ones, which has been calculated through Eq. \ref{eq10}
\begin{equation}
    \Delta SNR = |\frac{SNR_{denoised}-SNR_{noisy}}{SNR_{noisy}}| \cdot 100
    \label{eq10}
\end{equation}
It's crucial to notice the enhancement of SNR in the denoised images with respect to the noisy ones. In fact, denoised images present much higher SNR values, up to 61\% of increment.

\begin{table}[h!]
\centering
\caption[Linescan SNR values before and after denoising.]{Comparison of linescan SNR difference of denoised images SNR to noisy SNR values in percentage of noisy ones for SEM images with different underlayers (OUL, SOG), thicknesses (15, 30$\sim$nm), and frames of integration (4, 8, 16, 32, 64$\sim$Fr).}\label{tab:snr}
\vspace{3mm}
\begin{tabular}{|c|c|c|c|c|}
\cline{2-5}
\multicolumn{1}{c|}{} & \multicolumn{2}{c|}{\bf SOG} & \multicolumn{2}{c|}{\bf OUL} \\
\cline{2-5}
\multicolumn{1}{c|}{} & \bf 15$\sim$nm & \bf 30$\sim$nm & \bf 15$\sim$nm & \bf 30$\sim$nm \\
\hline
\bf 4 Fr & \color{red} 34.80 \% & 27.82	 \% & 33.58 \%	& 45.95  \% \\
\hline
\bf 8 Fr & 3.93 \% & 43.15 \%	& 56.66 \%	 & 47.90 \% \\
\hline
\bf 16 Fr & 27.78 \%	& 47.77 \%	& 61.06 \%	& 42.43 \% \\
\hline
\bf 32 Fr & \color{red} 174.31 \% &	42.16 \% &	53.62 \% &	34.36 \% \\
\hline
\bf 64 Fr & \color{red} 147.44 \% &	34.99 \% &	41.32 \% &	36.27 \%
 \\
\hline
\end{tabular}
\end{table}

The red values in Table \ref{tab:snr} are the only outliers, corresponding to the sample of SOG$\sim$15nm. Their presence can be explained easily by the \textit{SEM edge effect}. This phenomenon happens when patterning Lines/Spaces in correspondance of the edges of the lines. The secondary electrons, emitted by the material itself, are more in the edges than in the flat spaces of the pattern, therefore resulting in an increased brightness and higher contrast. Since the metric used to measure the SNR of the samples is correlated to the contrast and brightness, the values for noisier samples, such as SOG 15$\sim$nm, can easily result in outliers.

\subsection{Mean CD trend analysis}
\begin{figure} [h!]
\begin{center}
\begin{tabular}{c} %% tabular useful for creating an array of images 
\includegraphics[height=5cm]{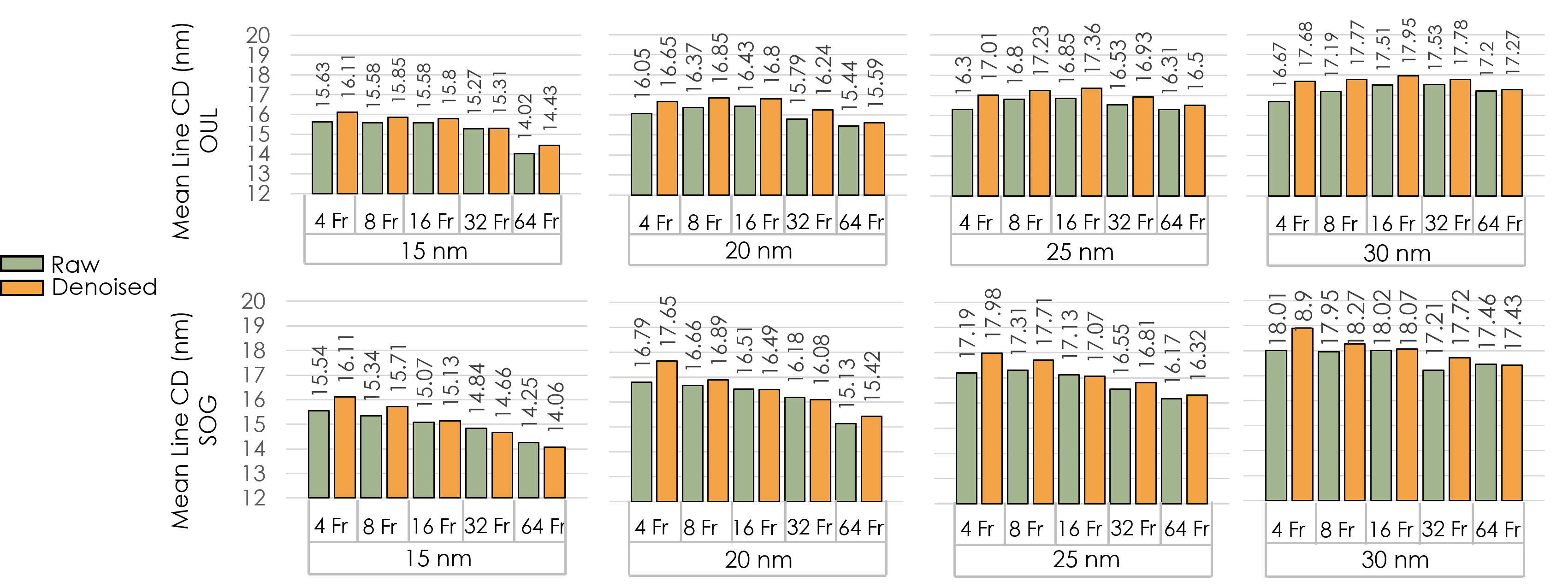}
\end{tabular}
\end{center}
\caption[Mean CD before and after denoising.]
%>>>> use \label inside caption to get Fig. number with \ref{}
{ \label{meancd} Mean CD before and after denoising for different underlayer (OUL, top graph; SOG, bottom graph), film thickness (15$\sim$nm, 20$\sim$nm, 25$\sim$nm, 30$\sim$nm) and frames of integration (4$\sim$Fr, 8$\sim$FR, 16$\sim$Fr, 32$\sim$Fr, 64$\sim$FR).}
\end{figure}

For each configuration of underlayer (SOG, OUL), film thickness (15$\sim$nm, 20$\sim$nm, 25$\sim$nm, 30$\sim$nm) and frames of integration (4$\sim$Fr, 8$\sim$Fr, 16$\sim$Fr, 32$\sim$Fr, 64$\sim$Fr), the mean CD of the raw/noisy SEM images, with target CD of 16$\sim$nm, has been investigated to understand if the process of denoising through the Machine Learning model has any repercussion on the detected CD. As shown in Fig. \ref{meancd}, the mean CD of denoised images is in full agreement with the raw data and it remains true to its trend.

Because of the nature of this experiment, there are no true ground truths for the CD of the tested devices, since it is a parameter affected by the SEM image acquisition settings, as well as multivariate process parameters. Moreover, the high number of frames of integration may cause the pattern shrinkage and, therefore, may change the absolute value of the target CD.

In Table \ref{tab000}, a comparison of the mean CD values before and after denoising is reported in percentage against noisy mean CD values, which have been calculated through Eq. \ref{eq11}
\begin{equation}
    \Delta CD = (\frac{CD_{noisy} - CD_{denoised}}{CD_{noisy}}) \cdot 100
    \label{eq11}
\end{equation}. For of the reasons explained above, it is not possible to define accurately a percentage threshold of acceptance for the difference in mean CD between noisy and denoised. However, to quantify as an experimental condition, mean CD values that are within a tolerance range of $\pm 5\%$ of noisy mean CD values (considered  as target) are considered compatible. This probes that the denoising procedure doesn't alter the L/S pattern dimensions.

\begin{table}[h!]
\centering
\caption{Comparison of mean CD difference of denoised images against noisy images (in percentage).}\label{tab000}
\vspace{3mm}
\begin{tabular}{|c|c|c|c|c|c|c|c|c|}
\cline{2-9}
\multicolumn{1}{c|}{} & \multicolumn{4}{c|}{\bf OUL} & \multicolumn{4}{c|}{\bf SOG} \\
\cline{2-9}
\multicolumn{1}{c|}{} & \bf 15$\sim$nm &  \bf 20$\sim$nm & \bf 25$\sim$nm & \bf 30$\sim$nm & \bf 15$\sim$nm & \bf 20$\sim$nm & \bf 25$\sim$nm & \bf 30$\sim$nm \\
\hline
\bf 4 Fr & \footnotesize -3.07 \%  & \footnotesize -3.74 \%  & \footnotesize -4.36 \%  & \footnotesize -6.06 \%  & \footnotesize -3.67 \% & \footnotesize -3.74 \% & \footnotesize -4.36 \% &  \footnotesize -6.06 \% \\
\hline
\bf 8 Fr & \footnotesize -1.73 \%  & \footnotesize -2.93 \%  & \footnotesize -2.56 \%  & \footnotesize -3.37 \%  &  \footnotesize -2.41 \%  & \footnotesize -2.93 \%  & \footnotesize -2.56 \%  &  \footnotesize -3.37 \%  \\
\hline
\bf 16 Fr & \footnotesize -1.41 \%  & \footnotesize -2.25 \%  & \footnotesize -3.03 \%  & \footnotesize -2.51 \%  &  \footnotesize -0.40 \%  & \footnotesize -2.25 \%  & \footnotesize -3.03 \%  &  \footnotesize -2.51 \%  \\
\hline
\bf 32 Fr & \footnotesize -0.26 \%  & \footnotesize -2.85 \%  & \footnotesize -2.42 \%  & \footnotesize -1.43 \%  & \footnotesize  1.21 \%  & \footnotesize -2.85 \%  & \footnotesize -2.42 \%  &  \footnotesize -1.43 \%  \\
\hline
\bf 64 Fr & \footnotesize -2.92 \%  & \footnotesize -0.97 \%  & \footnotesize -1.16 \%  & \footnotesize -0.41 \%  & \footnotesize  1.33 \%  & \footnotesize -0.97 \%  & \footnotesize -1.16 \%  & \footnotesize -0.41 \%   \\
\hline
\end{tabular}
\end{table}

\subsection{Roughness parameters}
\begin{figure}[htb!]
\centering
\captionsetup[subfigure]{}
    \subfloat[]{\includegraphics[width=7cm]{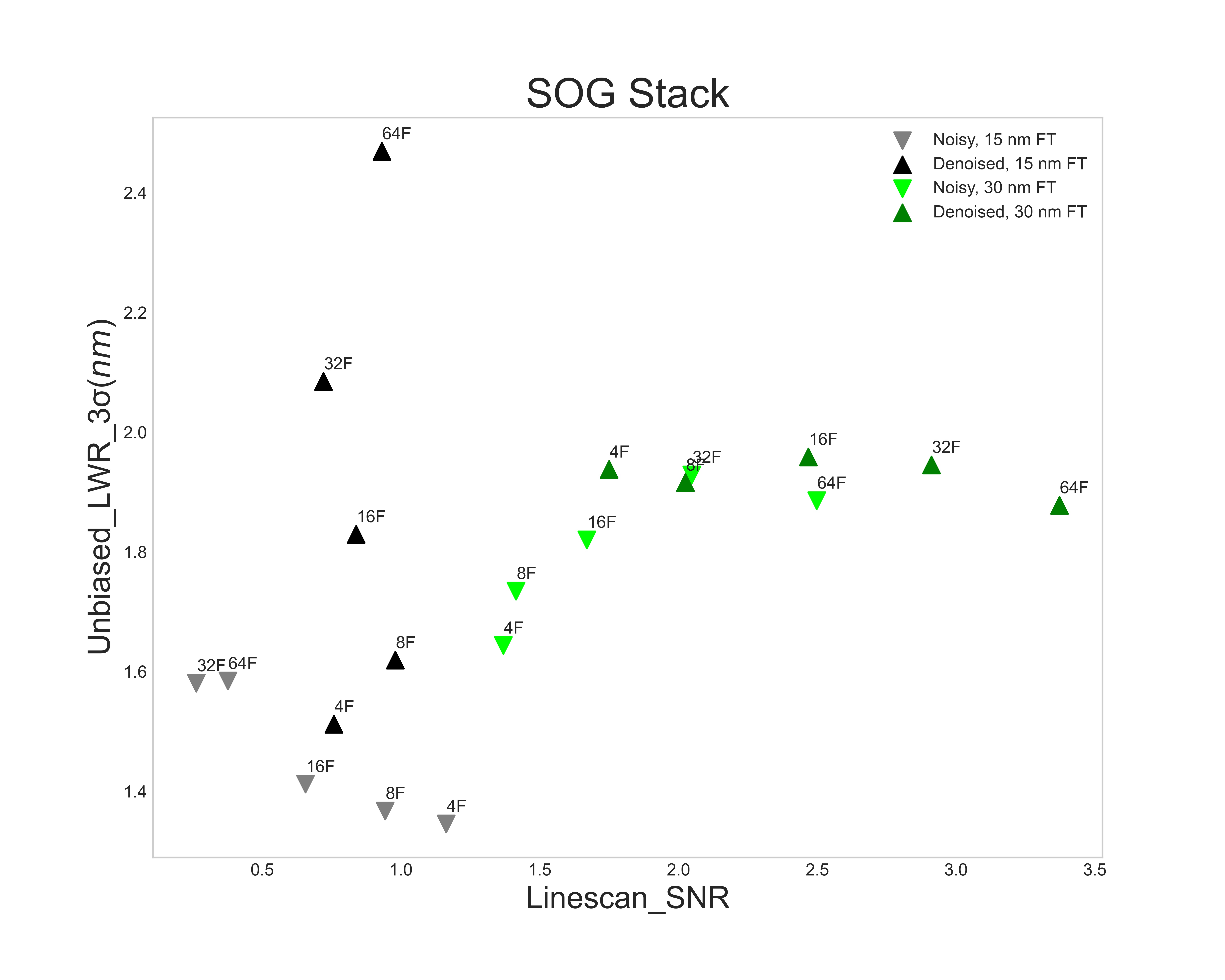} \label{SOG} }
    \qquad
    \subfloat[]{\includegraphics[width=7cm]{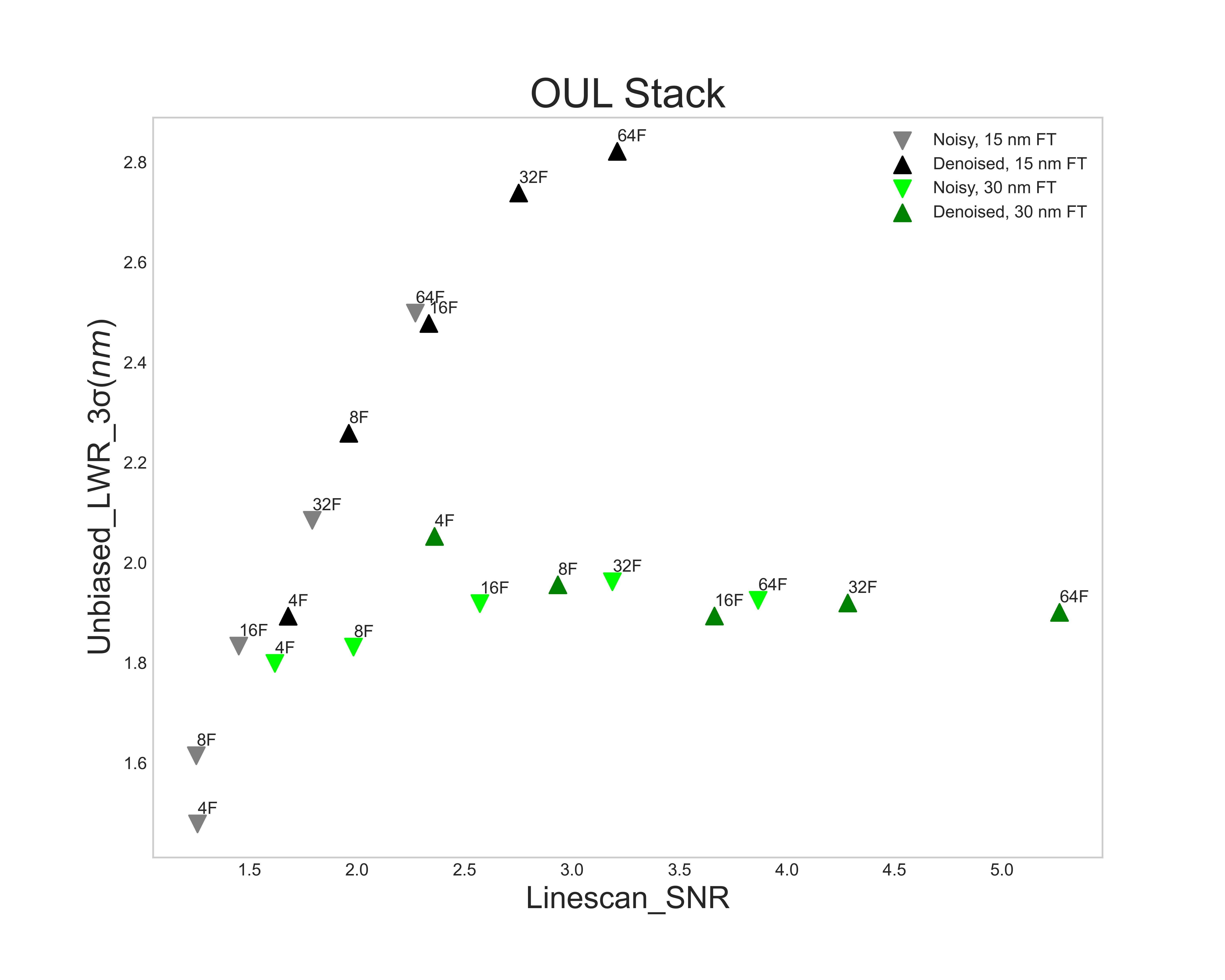} \label{OUL} }
    \caption{Comparison of uLWR against different line-scan SNR for raw and denoised images with different underlayers ((a) SOG and (b) OUL), film  thickness (15 and 30$\sim$nm), and frames of integration (4, 8, 16, 32, and 64 frames).}
    \label{snrlwr}
\end{figure}

As explained in the previous sections, the SNR has a significant impact on the unbiased LWR. Indeed, the reliability of the unbiasing procedure of the metrology software drops when the SNR$<2.0$. A necessary analysis consists in comparing the uLWR of raw and denoised images with the corresponding SNR, which can be seen in Fig. \ref{snrlwr}. These two scatter plots show the uLWR vs SNR trend for each underlayer, resist thickness, and number of frames. When comparing the two trends, raw and denoised, the first thing to notice is that the SNR values of denoised images are higher than those of the raw images. The uLWR of resists with 30$\sim$nm thickness for both underlayers follow the expected asymptotic plateu. In particular, for the OUL underlayer (Fig. \ref{OUL}) the uLWR values for the 15$\sim$nm thickness are greatly improved, almost reaching the asymptotic plateau. On the contrary, the uLWR values of the SOG underlayer (Fig. \ref{SOG}) do not follow the expected trend, since the corresponding SNR does not reach the threshold required for roughness measurements reliability (SNR $>2.0$), showing values around 1.0. 

\subsection{Partial Spectral Density (PSD) analysis}
\label{sec:psd}
\begin{figure} [h!]
\begin{center}
\begin{tabular}{c} %% tabular useful for creating an array of images 
\includegraphics[height=10cm]{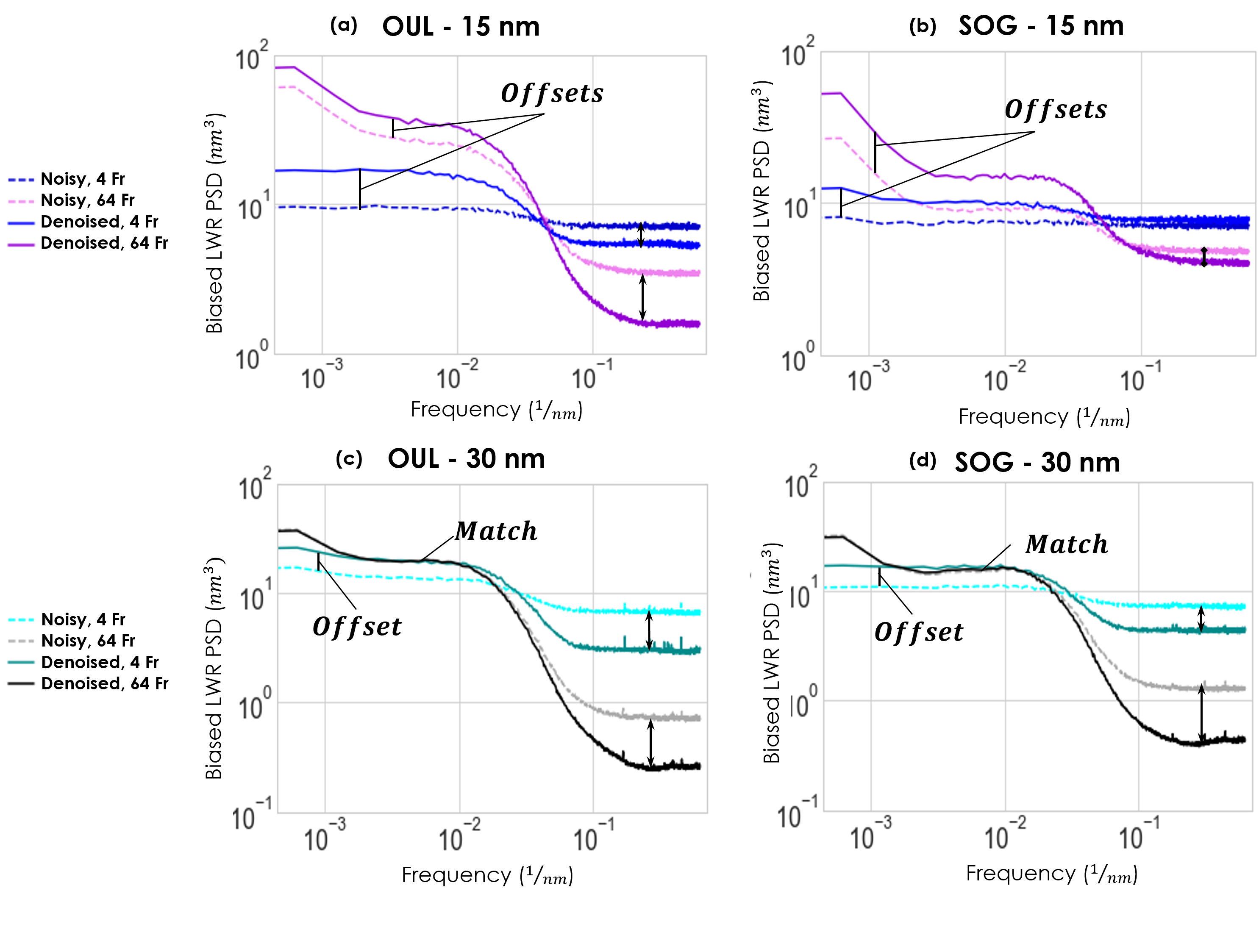}
\end{tabular}
\end{center}
\caption[Biased LWR PSD of SEM images.]
%>>>> use \label inside caption to get Fig. number with \ref{}
{ \label{biased} Biased LWR PSD of SEM images with (a) OUL underlayer, 15$\sim$nm thickness, (b) SOG underlayer, 15$\sim$nm thickness, (c) OUL underlayer, 30$\sim$nm thickness, and (d) SOG underlayer, 30$\sim$nm thickness.}
\end{figure}

One of the most relevant roughness metrics for L/S SEM images is Line Width Roughness (LWR). To conduct a thorough analysis of the noisy and denoised images for this experiment, biased and unbiased LWR  PSDs are extracted, plotted, and analyzed. In Fig. \ref{biased}, the biased LWR PSDs are shown.

Theoretically, biased LWR PSDs in the low frequency part of the graph should match for denoised and noisy, 4 and 64 frames, since the PSD trend in the low frequency region, called $PSD(0)$, characterizes the device features and needs to remain constant after denoising and when acquired with different frames of integration. Concurrently, the high frequency region of the PSD graph characterizes the acquisition noise, which should be significantly reduced after denoising the images.

It is relevant to notice that in the low frequency region of the graphs, for 15$\sim$nm of thickness, the PSD shows offsets between noisy and denoised curves, with the one for 4 frames being much larger than the one for 64 frames. This is expected, since the noisy images with 4 frames of integration have much more noise than with 64 frames, thus contributing to a larger offset. It can be derived that SOG as underlayer is more challenging than OUL, which can affect negatively the measurements even for 64 frames. Meanwhile, for the 30$\sim$nm of thickness there is an offset between the 4 frames noisy and denoised PSDs, while the 64 frames PSDs, noisy and denoised, match as expected. 

Instead, in the high frequency region, the biased LWR PSDs are following the expected trend, demonstrating the capability of the denoising model, since the denoised images exhibit significantly lower values of noise than the raw ones. 

To summarize, the biased and unbiased LWR PSDs of denoised images follow the expected trend by being lowered in the high frequency region. The only exception is for SOG with 15 nm of thickness with 4 frames, in which the PSDs of both noisy and denoised images almost overlap. The offsets in the low frequency region are dependent on the PSD computation carried out in the background by the metrology software itself, which can't be improved by the denoising procedure.

It is essential to highlight how the LWR PSDs in low frequency region for 30$\sim$nm of thickness are not presenting any offset for 64 frames (e.g., matching) and comparatively lesser one for 4 frames, thus, they are consistently better than those of 15$\sim$nm of thickness. This is also expected, since the CD-SEM noise is known to be increased for thinner photoresists.

\begin{figure}
\centering
\captionsetup[subfigure]{labelformat=empty}
    \subfloat[]{\includegraphics[width=12cm]{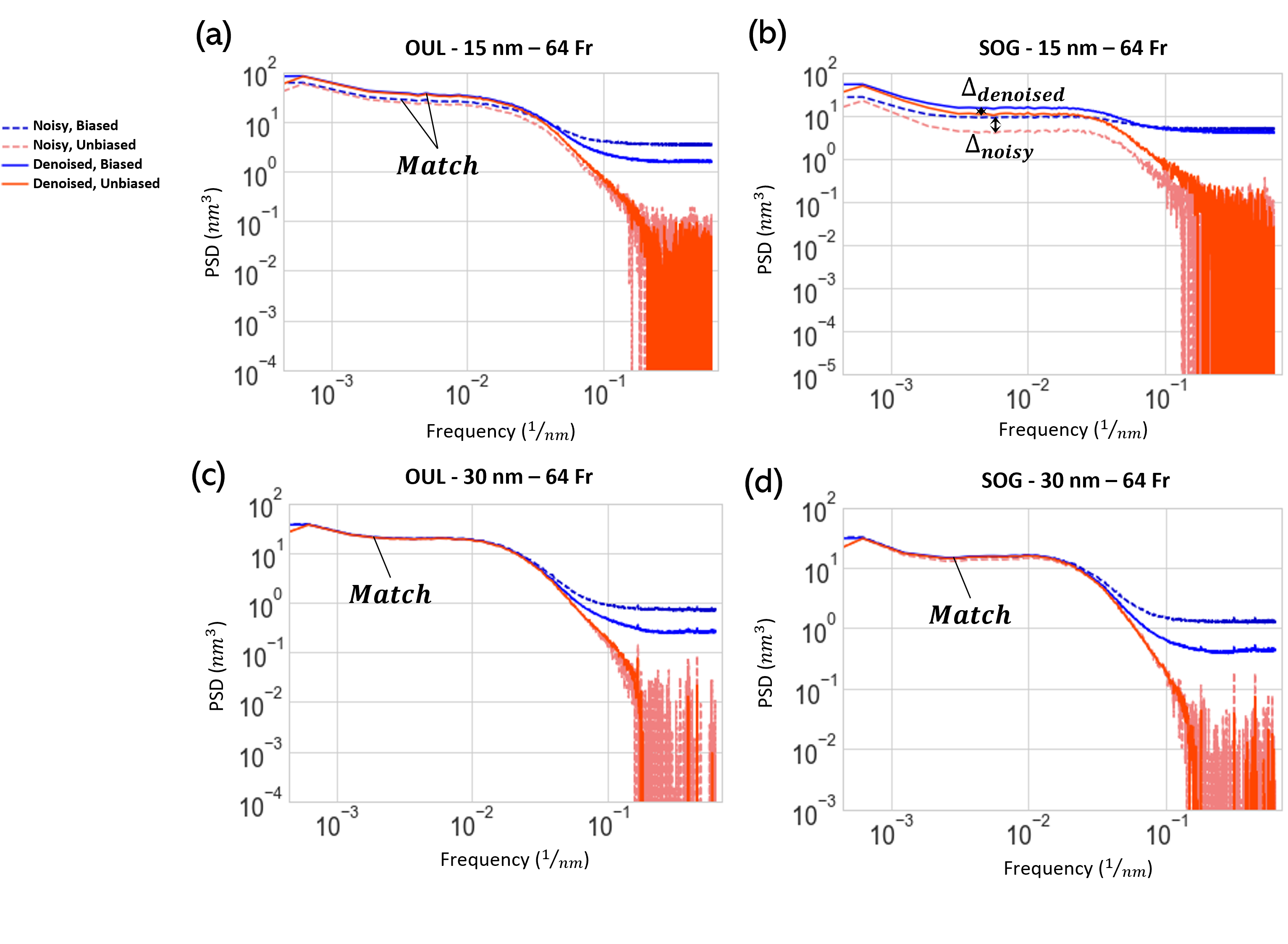} \label{comp64} }
    \qquad
    \subfloat[]{\includegraphics[width=12cm]{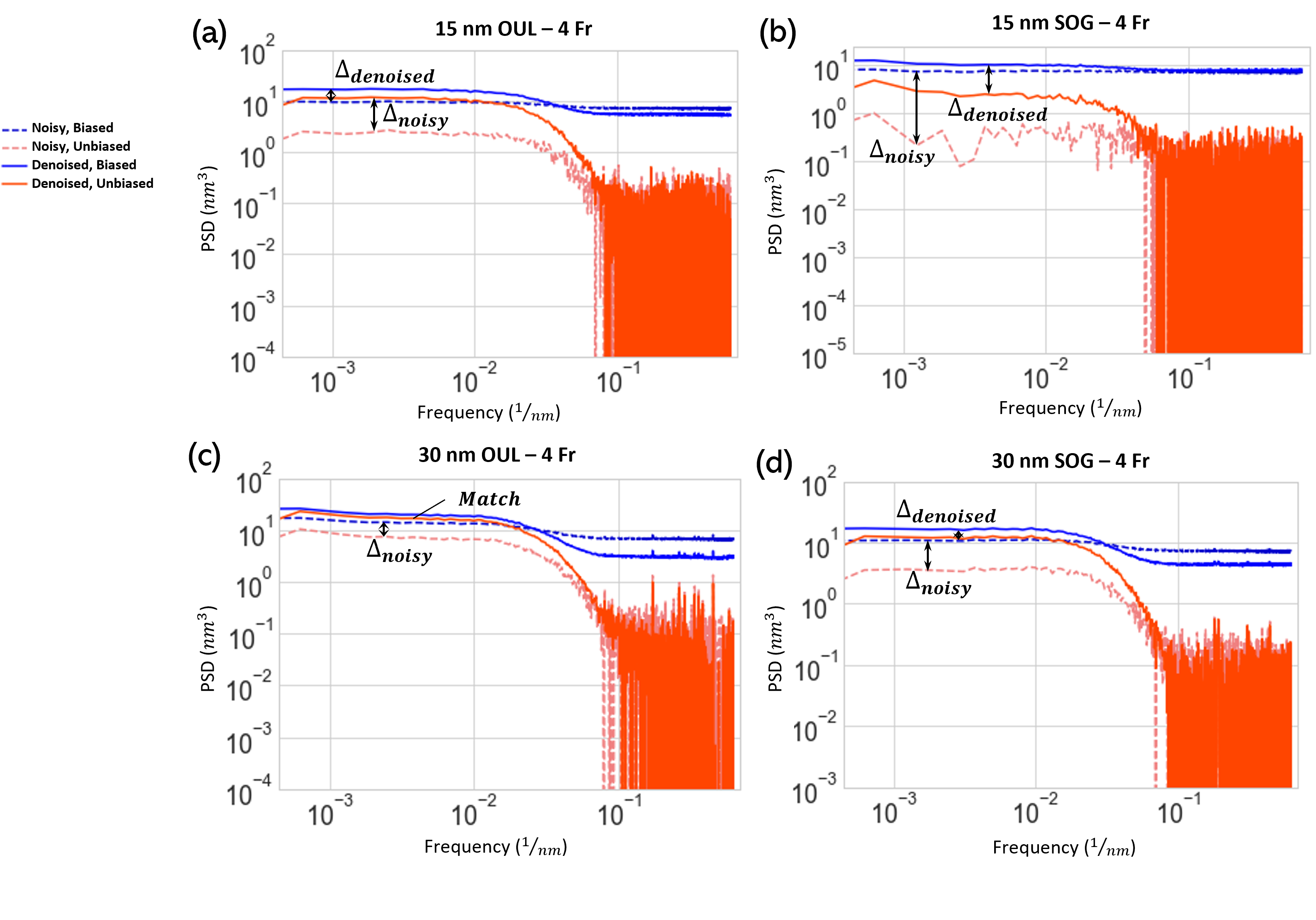} \label{comp4} }
    \caption[Comparison of biased and unbiased LWR PSD.]{Comparison of biased and unbiased LWR PSDs of SEM images for 64 (top graph) and 4 frames (bottom graph) with (a) OUL underlayer, 15$\sim$nm thickness, (b) SOG underlayer, 15$\sim$nm thickness, (c) OUL underlayer, 30$\sim$nm thickness, and (d) SOG underlayer, 30$\sim$nm thickness.}
    \label{comp}
\end{figure}

To conclude, the comparison between the biased and unbiased LWR PSDs are shown in Fig. \ref{comp} for both SOG and OUL underlayers, 15 and 30$\sim$nm of thickness, 4 and 64 frames of integration. In these plots the offsets between biased and unbiased noisy and denoised PSDs are indicated with $\Delta_{noisy}$, and $\Delta_{denoised}$ respectively. It is relevant to notice that, as before, there is a fundamental difference between the 15$\sim$nm and 30$\sim$nm PSDs, given by the much higher SEM noise of thinner photoresists. Moreover, the comparison between the two different underlayers shows how SOG is more challenging than OUL, resulting in much larger $\Delta_{noisy}$ and $\Delta_{denoised}$ for both 15 and 30$\sim$nm, 4 and 64 frames settings.
At last, an important result lies in the difference between 4 and 64 frames. In Fig. \ref{comp} for 64 Fr (top graph), the PSDs show an exact matching in the 30$\sim$nm samples for both OUL and SOG, and an almost exact in OUL 15$\sim$nm. SOG 15$\sim$nm sample is the only one presenting very small $\Delta_{noisy}$ and $\Delta_{denoised}$ offsets (being $\Delta_{denoised}$ itself smaller than $\Delta_{noisy}$).
On the contrary, graphs for 4 Fr shows perceptible offsets, both for noisy and denoised images, which is expected because of the computation by SMILE of the unbiased PSDs, which in turn adds inevitably an irreducible offset. A most significant fact to be noticed here is, the $\Delta_{denoised}$ for all conditions are marginal (or matched) in comparison to corresponding $\Delta_{noisy}$.

It's ultimately important to underline that the denoising procedure for images with 4 Fr, which have lower SNR than 64 Fr, is capable of obtaining PSD analysis comparable with that of 64 Fr. This result indicates that by denoising SEM images acquired with low frames of integration using a DL based denoiser \cite{dey2021sem} it is now possible to extract reliable roughness measurements, such as LWR and LER, while saving time in the production process and decreasing the risk of damaging the printed patterns.

\clearpage

\section{CONCLUSIONS}
As research on High NA EUVL for high volume manufacturing is stepping up, there is more and more necessity of reliable metrology for the thin resists employed in this technique, which is impacted by the reduced imaging contrast and low SNR resulting from the low thickness of resists (below 30$\sim$nm). SEM images acquired with a larger number of frames tend to be less noisy but require more (tool and engineering) time and cost, while those with a smaller number of frames are timely more efficient, but result in a lower SNR, which impacts the reliability of the unbiased LWR and LER. 
In order to enhance the SNR, a Deep Learning denoiser has been employed on SEM images of different underlayers (SOG, OUL), film thickness (15, 20, 25, 30$\sim$nm), and frames of integration (4, 8, 16, 32, 64 Fr). 
Firstly, the denoised images exhibit an enhanced SNR, up to an increment in the linescan SNR values of 61\%. Secondly, it has been established that the mean CD values of the SEM images are not altered by the denoising procedure, in fact their percentual difference lies under the threshold of 5\% for almost all combinations. Moreover, the analysis on the LER/LWR PSDs conducted through an open-source metrology software, SMILE 2.3.2, shows that, in the high frequency region of the biased LWR PSD the noise of the images is significantly reduced by the denoising procedure. Meanwhile, in the low frequency region of the biased LWR PSD, the PSD curves match (before/after denoising) when the SNR is higher than 2.0, while offsets between noisy and denoised PSDs are observed in most demonstrations with reduced (15 nm) resist thickness and/or smaller (4 Fr) number of frames, due to the higher SEM noise of images acquired from samples with thinner photoresists ($SNR < 2$).
In general, it can be assessed that the LER/LWR PSD analysis for denoised images acquired with 4 frames of integration has a compatible accuracy and reliability with respect to noisy images acquired with 64 frames. To conclude, denoising SEM images with low number of frames using Machine Learning has proven to be helpful in preserving the roughness measurements reliability for High NA EUVL thin resists and increasing the metrology throughput.

%\acknowledgments % equivalent to \section*{ACKNOWLEDGMENTS} 
\acknowledgements
The authors used the imec, Leuven, research computing facility to conduct part of the research. We thank Mohamed Zidan for providing the experimental dataset.

% References
\bibliography{report} % bibliography data in report.bib
\bibliographystyle{spiebib} % makes bibtex use spiebib.bst

\end{document}